\documentclass[runningheads,a4paper]{llncs}

\usepackage[T1]{fontenc}

\usepackage{amsmath,amssymb}

\usepackage{algorithm}
\usepackage{algorithmic}

\usepackage{graphicx}
\usepackage{textcomp}
\usepackage{booktabs}
\usepackage{multirow}

\usepackage{tikz}
\usetikzlibrary{shapes.geometric, arrows.meta, positioning, patterns, calc}
\usepackage{pgfplots}
\pgfplotsset{compat=1.18}
\usepgfplotslibrary{fillbetween}
\usepgfplotslibrary{statistics}

\definecolor{oiblue}{HTML}{0072B2}       % Primary: active/processing strokes
\definecolor{oiorange}{HTML}{E69F00}     % Accent: final/selected/important
\definecolor{oiskyblue}{HTML}{56B4E9}    % Light: backgrounds/fills
\definecolor{oigray}{HTML}{999999}       % Neutral: inactive/pruned
\definecolor{oivermillion}{HTML}{D55E00} % Warning/emphasis (replaces red)
\definecolor{oigreen}{HTML}{009E73}      % Alternative accent (bluish-green)
\definecolor{oipurple}{HTML}{CC79A7}     % Secondary accent (reddish-purple)

\usepackage[hidelinks,hypertexnames=false]{hyperref}
\hypersetup{
  pdftitle={On Scaling Coordinate-Based Neuroevolution: The Quadtree Bottleneck in ES-HyperNEAT},
  pdfauthor={Romain Claret, Michael O'Neill, Paul Cotofrei, Kilian Stoffel},
  pdfsubject={Neuroevolution; ES-HyperNEAT; GPU parallelization},
  pdfkeywords={ES-HyperNEAT, Neuroevolution, Adaptive substrate evolution, CPPN, GPU parallelization, Vectorization barriers, Quadtree}
}

\begin{document}

\title{On Scaling Coordinate-Based Neuroevolution:\\
The Quadtree Bottleneck in ES-HyperNEAT\thanks{Preprint. This is an extended version of a manuscript originally prepared for double-blind review, released as a
standing diagnostic companion to the EMR-HyperNEAT work~\cite{claret2026emr} that resolves the
bottleneck characterized here (Section~\ref{sec:emr}).}}
\titlerunning{On Scaling Coordinate-Based Neuroevolution}

\author{Romain Claret\inst{1} \and Michael O'Neill\inst{2} \and Paul Cotofrei\inst{1} \and Kilian Stoffel\inst{1}}
\authorrunning{R. Claret et al.}
\institute{University of Neuch\^{a}tel, Information Management Institute, Neuch\^{a}tel, Switzerland\\
\email{\{romain.claret,paul.cotofrei,kilian.stoffel\}@unine.ch}
\and University College Dublin, Natural Computing Research \& Applications Group, Dublin, Ireland\\
\email{m.oneill@ucd.ie}}

\maketitle

\begin{abstract}
ES-Hyper\-NEAT evolves substrate topology through adaptive quad\-tree subdivision; to our knowledge no im\-ple\-men\-ta\-tion with full population-level GPU par\-al\-lel\-iza\-tion exists. We present JAX-ESHN, a JAX-based im\-ple\-men\-ta\-tion targeting GPU par\-al\-lel\-iza\-tion with batched CPPN queries, and benchmark it against the CPU-based PUREPLES Baseline across five tasks: XOR, Parity-3, circle clas\-si\-fi\-ca\-tion, sine regression, and CartPole. The core limitation is structural: each CPPN discovers a unique set of substrate positions, preventing population-level vec\-tor\-iza\-tion via \texttt{vmap}. On XOR, the CPU Baseline's runtime scales exponentially with depth while JAX-ESHN's construction cost on GPU (compilation plus first-generation evaluation) plateaus at deep substrates, so JAX-ESHN solves reliably where the Baseline rarely succeeds, with lower runtime variance. A CPU-vs-CPU multi-benchmark control reproduces the same scaling divergence across Boolean, continuous, and control task types, confirming it is a property of the substrate-discovery im\-ple\-men\-ta\-tion, not of GPU hardware. An alternative data structure (Hierarchical Spatial Hash Grid) fails not because it precomputes positions but because it applies the variance test independently per position, discarding the quadtree's parent-gated filtering and with it the adaptive sparsity essential to ES-HyperNEAT. These findings define the structural constraints any substrate-discovery method must satisfy to scale coordinate-based neuroevolution; the companion EMR-HyperNEAT re\-for\-mu\-la\-tion, which replaces adaptive subdivision with eager evaluation of a static multi-resolution grid, satisfies them and resolves the bottleneck this paper characterizes.

\keywords{ES-HyperNEAT \and Neuroevolution \and Adaptive substrate evolution \and CPPN \and GPU parallelization \and Vectorization barriers \and Quadtree}
\end{abstract}

\section{Introduction}
\label{sec:introduction}

Evolvable-Substrate HyperNEAT (ES-HyperNEAT)~\cite{risi2012enhanced} extends HyperNEAT by evolving substrate topology through adaptive quadtree subdivision. While standard HyperNEAT~\cite{stanley2009hypercube} requires predefined substrate architectures, ES-Hy\-per\-NEAT evolves the substrate topology itself, automatically discovering where neurons should be placed based on information density in the Compositional Pattern Producing Network (CPPN)~\cite{stanley2007compositional}. This lets networks discover task-appropriate topologies.

Adaptive quadtree subdivision is itself a self-organizing process: substrate topology emerges from a local variance-driven subdivision rule applied to a CPPN-generated field, not from any predetermined template. The reference implementation (PUREPLES~\cite{westh2017pureples}), referred to as the \textit{Baseline} throughout, runs on CPU with sequential quadtree traversal per individual. Evolutionary algorithms require large populations evaluated over many generations, and the position count grows exponentially with substrate depth, making GPU acceleration valuable. JAX~\cite{bradbury2018jax}, with just-in-time (JIT) compilation, and TensorNEAT~\cite{wang2024tensorized} provide GPU-accelerated NEAT and HyperNEAT, raising the question of whether ES-HyperNEAT can similarly benefit.

We hypothesize that ES-HyperNEAT's quadtree presents a structural bottleneck that optimization cannot overcome. The core issue is that each CPPN in the population discovers a \textit{unique} set of substrate positions through quadtree subdivision. This per-CPPN variability prevents population-level vectorization: JAX's \texttt{vmap} requires identical array shapes across batch elements, but variable-length quadtree outputs differ for each CPPN. This limits GPU utilization for any implementation built on static-shape compilation (detailed in Section~\ref{sec:optimizations}). The finding exposes a tension in \emph{adaptive} indirect encoding: the mechanism that lets each genome self-organize a sparse, problem-tailored substrate is the same one that breaks the uniformity assumption population-level GPU batching relies on. Adaptivity and static-shape vectorizability are therefore in direct opposition for any adaptive substrate-discovery method that produces variable-cardinality position sets, the class this paper diagnoses.

We present \textit{JAX-ESHN} (JAX-based ES-HyperNEAT), an implementation targeting GPU parallelization using TensorNEAT~\cite{wang2024tensorized} for CPPN evolution and JAX for substrate construction. This work investigates three research questions: (RQ1) Can batched optimizations overcome the sequential nature of quadtree-based substrate discovery? (RQ2) How does JAX-ESHN compare to the Baseline in solve rate and efficiency? (RQ3) What is the scalability limit of quadtree-based substrate discovery?

\subsection{Contributions}

This paper makes four contributions. We give mechanistic and empirical evidence that ES-Hyper\-NEAT's adaptive quadtree is structurally incompatible with static-shape compilation frameworks (JAX \texttt{vmap}/XLA, TensorFlow static graphs, and other JIT-traced systems), because position discovery runs per CPPN. Dynamic-shape primitives such as hand-written CUDA kernels and rag\-ged-batch GPU runtimes remain a theoretical escape, which we assess without ruling out in Section~\ref{sec:discussion}. JAX-ESHN itself reaches near-100\% solve rates at depths where the Baseline degrades to 4.2\%, with faster solve times and generations-to-solve that fall with depth instead of rising (Section~\ref{subsec:construction-cost}). The speed advantage comes from solving during the construction generation rather than from amortizing it, and the cross-implementation solve-rate comparison carries a NEAT-library confound (Section~\ref{sec:caveats}). The Hierarchical Spatial Hash Grid (HSHG) then shows why a fixed grid fails when variance is tested per position: par\-ent-gat\-ing is lost, and adaptive sparsity with it, causing order-of-mag\-ni\-tude over-dis\-cov\-ery. Finally, ANOVA across 7 depths, 8 populations, and 456 trials identifies depth as the dominant factor, and a CPU-vs-CPU control extending beyond XOR to four further tasks (Parity-3, circle classification, sine regression, and CartPole, spanning Boolean, continuous, and control problems) shows the scaling divergence persists across task types and on CPU alone.

The five benchmarks studied here (XOR, Parity-3, circle, sine, CartPole) are deliberately chosen for their cost-controlled scaling behavior rather than for representational depth; extending this analysis to higher-dimensional benchmarks such as MNIST is a natural follow-on. The known scaling limitations of HyperNEAT on complex domains do not weaken the case for diagnosing this bottleneck: on ES-HyperNEAT, hyperparameter optimization~\cite{claret2024investigating}, trial-level early stopping~\cite{claret2026pruner}, and spatial input partitioning~\cite{claret2026partitioned} all still benefit from GPU acceleration even at toy scale, and the diagnostic mechanism (per-CPPN variable position discovery $\rightarrow$ static-shape vectorization barrier) recurs in any adaptive-substrate method that pre-discovers variable-cardinality position sets. A clear negative result also has value here: confirming a structural ceiling prevents the field from repeatedly investing engineering effort in implementations doomed by the same root cause. The remainder of this paper is organized as follows. Section~\ref{sec:background} reviews ES-HyperNEAT and GPU-accelerated neuroevolution. Section~\ref{sec:implementation} describes our implementation architecture. Section~\ref{sec:optimizations} presents four batched optimizations and their limitations. Section~\ref{sec:hshg} tests an alternative data structure. Section~\ref{sec:results} provides experimental results across five benchmarks. Section~\ref{sec:discussion} discusses implications and practical guidance. Section~\ref{sec:conclusion} concludes.

\section{Background and Related Work}
\label{sec:background}

We review the neuroevolutionary techniques that ES-HyperNEAT builds upon and examine existing implementations.

\subsection{NEAT, HyperNEAT, and ES-HyperNEAT}

NEAT (NeuroEvolution of Augmenting Topologies)~\cite{stanley2002evolving} evolves neural network topologies and connection weights through in\-no\-va\-tion tracking, speciation, and complexification from minimal structures. HyperNEAT~\cite{stanley2009hypercube} uses NEAT to evolve CPPNs that generate connection weights for a predefined substrate based on spatial coordinates, $w = \text{CPPN}(x_1, y_1, x_2, y_2)$. This indirect encoding exploits geometric regularities, so compact genomes can generate large-scale structured networks~\cite{clune2011performance}.

ES-HyperNEAT~\cite{risi2012enhanced} uses adaptive quadtree subdivision~\cite{finkel1974quad} to evolve substrate topology. Regions where the CPPN produces high output variance subdivide further; uniform regions terminate early. This variance-guided discovery means each CPPN generates a unique set of substrate positions. However, this adaptive subdivision incurs computational costs that grow exponentially with depth, becoming prohibitive beyond depth 5~\cite{claret2024investigating}.

Three lines of prior work respond to this cost from outside the algorithm. Systematic hyperparameter optimization identifies configurations ES-HyperNEAT can exploit~\cite{claret2024investigating}; a data-driven early-stopping rule prunes unpromising trials from that search, cutting its budget by 41.6\% in evaluated generations on MNIST without degrading the solutions returned~\cite{claret2026pruner}; and partitioning a high-dimensional input across separately evolved specialists lifts mean accuracy on the same benchmark from 20.95\% to 43.07\% by keeping each run within the input dimensionality a fixed generation budget can exploit~\cite{claret2026partitioned}. Each makes a search over ES-HyperNEAT cheaper or more effective, and none alters the substrate-discovery procedure itself: the sequential quadtree traversal that turns one CPPN into one substrate, and its exponential growth with depth, survives all three. Reducing that per-evaluation cost requires working on the algorithm itself, which is what this paper examines.

The Baseline uses neat-python~\cite{neatpython2015} for CPPN evolution. TensorNEAT~\cite{wang2024tensorized} implements NEAT in JAX~\cite{bradbury2018jax} for GPU acceleration through XLA (Accelerated Linear Algebra) compilation~\cite{sabne2020xla}. JAX-ESHN builds upon TensorNEAT's CPPN evolution infrastructure.

\subsection{GPU-Accelerated Neuroevolution}

JAX~\cite{bradbury2018jax}'s \texttt{jit} traces Python functions into XLA graphs with \emph{fixed} tensor shapes; \texttt{vmap} maps functions over batch dimensions, requiring \emph{uniform} shapes across all elements. This uniform-shape requirement is central to the incompatibility we identify. Frameworks like TensorNEAT~\cite{wang2024tensorized}, EvoJAX~\cite{tang2022evojax}, evosax~\cite{lange2023evosax}, and QDax~\cite{lim2022qdax} succeed because they evaluate \emph{fixed-architecture} networks across populations; ES-HyperNEAT breaks this assumption because each CPPN discovers a different substrate topology, producing variable-shape networks that cannot be batched. This is also why we report no head-to-head against those GPU frameworks: each requires a fixed substrate, so none discovers the variable-cardinality topologies whose batching is the question, and the absence of any comparable variable-substrate GPU system is itself the finding. TensorNEAT, the one JAX framework that evolves network topology, implements NEAT and HyperNEAT but not ES-HyperNEAT~\cite{wang2024tensorized}; to our knowledge, no GPU-accelerated ES-HyperNEAT existed before this work. Precomputed-grid or approximate-quadtree substitutes are conceivable, but any scheme that pre-generates candidate positions and filters them independently inherits the over-discovery we document in Section~\ref{sec:hshg}.

\section{Implementation Architecture}
\label{sec:implementation}

JAX-ESHN operates as a two-level evolutionary system: NEAT (via Ten\-sor\-NEAT) evolves CPPN genomes, while JAX-based substrate discovery builds and evaluates the resulting networks. JAX-ESHN's substrate discovery follows the same adaptive quadtree algorithm as the Baseline. We describe the generation loop, three-phase substrate discovery, and the division and pruning algorithms below.

\subsection{System Overview}

Throughout this paper, ``depth'' refers to \texttt{max\_depth} in the implementation: the maximum quadtree level including the initial subdivision; depth $d$ yields $\frac{4^{d+1} - 1}{3}$ potential positions. The substrate uses 2D coordinates $(x, y) \in [-1, 1]^2$, with $y$ representing layer position, and a feedforward constraint on $y$ orders the connections. The two implementations apply it slightly differently ($s.y < \ell.y$ in the Baseline, $s.y \le \ell.y$ in JAX-ESHN); Section~\ref{sec:caveats} takes this up.

\begin{figure}[!b]
\centering
\resizebox{\textwidth}{!}{%
\begin{tikzpicture}[scale=0.72, every node/.style={scale=0.72},
    box/.style={rectangle, draw, thick, minimum width=2cm, minimum height=0.8cm, align=center},
    arrow/.style={-{Stealth}, thick}
]
    \node[box, fill=oiblue!15] (neat) at (0,2) {TensorNEAT\\(NEAT)};
    \node[box, fill=oigreen!15] (substrate) at (3.5,2) {Substrate\\Discovery};
    \node[box, fill=oigray!15, minimum width=1.5cm] (cppn) at (0,0) {CPPN\\Genomes};
    \node[box, fill=oigray!15, minimum width=1.5cm] (nets) at (3.5,0) {Substrate\\Networks};
    \node[box, fill=oiorange!15, minimum width=1.5cm] (fit) at (6.5,1) {Fitness\\Eval};
    \draw[arrow] (neat) -- (substrate) node[midway, above, font=\small] {CPPNs};
    \draw[{Stealth}-{Stealth}, thick] (neat) -- (cppn);
    \draw[arrow] (substrate) -- (nets);
    \draw[arrow] (nets) -- (fit);
    \draw[arrow] (fit) -- ++(0,2.5) -| (neat) node[pos=0.25, above, font=\small] {Fitness Score};
    \draw[dashed, thick, rounded corners] (-1.2,-0.8) rectangle (5.2,3.3);
    \node[font=\small, anchor=north east] at (5.1,3.2) {Per-generation loop};
    \node[font=\scriptsize, text=oivermillion] at (3.5,-1.2) {Sequential (cannot vmap)};
\end{tikzpicture}}
\caption{JAX-ESHN system architecture. TensorNEAT evolves CPPN genomes (left); each is processed by substrate discovery (center) to build a network via quadtree subdivision; fitness scores (right) drive selection. The bottleneck: substrate discovery runs sequentially per CPPN, between the batched \texttt{ask} and \texttt{tell}, because each CPPN's topology differs (Section~\ref{sec:introduction}).}
\label{fig:architecture}
\end{figure}
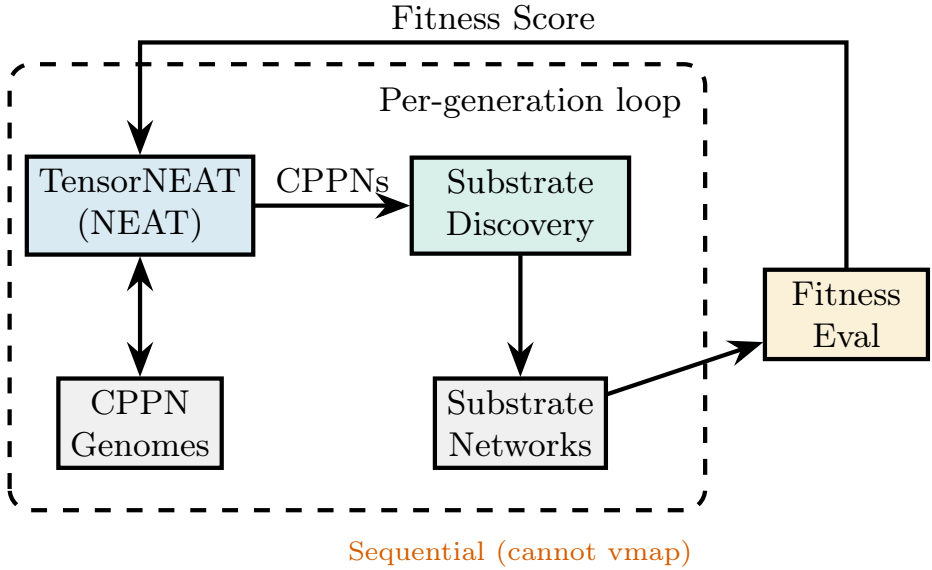

JAX-ESHN's generation loop follows the standard NEAT pattern. Ten\-sor\-NEAT returns the population, and a population-level \texttt{vmap}(transform) normalizes CPPN genomes to weight matrices. Each CPPN is then processed sequentially (substrate discovery, network construction, fitness evaluation), and the fitness vector is returned to TensorNEAT for selection. The per-CPPN inner loop (Algorithm~\ref{alg:system}, lines~3--8) cannot be vectorized due to variable topology discovery (Section~\ref{sec:introduction}). Figure~\ref{fig:architecture} shows where that sequential work sits between the batched \texttt{ask} and \texttt{tell}.

\begin{algorithm}[t]
\caption{ES-HyperNEAT Generation Loop}
\label{alg:system}
\begin{algorithmic}[1]
\REQUIRE State $s$, Problem $P$, Population size $N$
\STATE $\text{cppns} \leftarrow \text{TensorNEAT.ask}(s)$ \COMMENT{Sample CPPN population}
\STATE $\text{cppns\_transformed} \leftarrow \text{vmap}(\text{transform})(s, \text{cppns})$
\FOR{$i = 1$ \TO $N$}
    \STATE $\text{cppn} \leftarrow \text{cppns\_transformed}[i]$
    \STATE $H, C \leftarrow \textsc{DiscoverSubstrate}(\text{cppn})$
    \STATE $\text{net} \leftarrow \textsc{BuildNetwork}(H, C)$
    \STATE $f_i \leftarrow \textsc{Evaluate}(\text{net}, P)$
\ENDFOR
\STATE $s' \leftarrow \text{TensorNEAT.tell}(s, \mathbf{f})$
\RETURN $s'$
\end{algorithmic}
\end{algorithm}

\subsection{Three-Phase Substrate Discovery}

Substrate discovery proceeds in three phases because connectivity depends on previously discovered nodes.

\begin{itemize}
    \item \emph{Phase~1.} Finds input$\to$hidden connections.
    \item \emph{Phase~2.} Discovers hidden$\to$hidden connections over $L$ iterations, set by \texttt{iteration\_level}; $L=0$ yields single-layer networks. Both implementations run $L=1$ here, so Phase~2 performs one expansion round: each Phase~1 hidden node is explored outward, adding hid\-den-to-hid\-den connections and any further nodes they reach.
    \item \emph{Phase~3.} Connects hidden$\to$output.
\end{itemize}

Each phase constructs a quadtree using variance-driven subdivision, then extracts connections via band-detection pruning.

\subsection{Division Initialization (Quadtree Subdivision)}

Division initialization uses two depth parameters: $D_{\text{init}}$ forces subdivision to a minimum depth regardless of variance, while $D_{\text{max}}$ sets the hard ceiling for subdivision (\texttt{initial\_depth} and \texttt{max\_depth} in the configurations). Algorithm~\ref{alg:division} states the procedure; its variance-gated subdivision branch is the exact step that makes the output cardinality CPPN-dependent and thus breaks static-shape batching.

\begin{algorithm}[t]
\caption{Division Initialization (quadtree subdivision)}
\label{alg:division}
\begin{algorithmic}[1]
\REQUIRE CPPN $\theta$, Source coord $s$, Direction flag $d$
\REQUIRE $D_{\text{init}}$, $D_{\text{max}}$, $\tau_{\text{div}}$
\STATE $\text{root} \leftarrow \text{QuadPoint}(0, 0, 1.0, \text{level}=1)$
\STATE $Q \leftarrow [\text{root}]$ \COMMENT{BFS queue}
\WHILE{$Q \neq \emptyset$}
    \STATE $p \leftarrow Q.\text{pop}()$;\quad $w \leftarrow p.\text{width} / 2$
    \STATE $\text{children} \leftarrow$ 4 QuadPoints at $(p.x \pm w, p.y \pm w)$, level $p.\text{level}{+}1$
    \FOR{each $c \in \text{children}$}
        \STATE $c.\text{weight} \leftarrow \textsc{QueryCPPN}(\theta, s, (c.x, c.y), d)$
    \ENDFOR
    \STATE $\sigma^2 \leftarrow \text{Var}([\text{children}[i].\text{weight}])$
    \IF{$p.\text{level} < D_{\text{init}}$ \OR ($p.\text{level} < D_{\text{max}}$ \AND $\sigma^2 > \tau_{\text{div}}$)}
        \STATE $p.\text{children} \leftarrow \text{children}$;\quad $Q.\text{extend}(\text{children})$
    \ENDIF
\ENDWHILE
\RETURN root
\end{algorithmic}
\end{algorithm}

\begin{figure}[t]
\centering
\begin{tikzpicture}[scale=0.72]
    % Initial root (left)
    \draw[thick] (0,0) rectangle (2,2);
    \node at (1,1) {Root};
    \node[below] at (1,0) {\small Level 1};

    % Arrow
    \draw[-{Stealth}, thick] (2.3,1) -- (3.2,1);
    \node[above] at (2.75,1) {\small $\sigma^2 > \tau$};

    % First subdivision (middle)
    \draw[thick] (3.5,0) rectangle (5.5,2);
    \draw[thick] (4.5,0) -- (4.5,2);
    \draw[thick] (3.5,1) -- (5.5,1);
    \node[font=\tiny] at (4,1.5) {TL};
    \node[font=\tiny] at (5,1.5) {TR};
    \node[font=\tiny] at (4,0.5) {BL};
    \node[font=\tiny] at (5,0.5) {BR};
    \node[below] at (4.5,0) {\small Level 2};

    % Arrow
    \draw[-{Stealth}, thick] (5.8,1) -- (6.7,1);

    % Final subdivision (right): one quadrant further subdivided
    \draw[thick] (7,0) rectangle (9,2);
    \draw[thick] (8,0) -- (8,2);
    \draw[thick] (7,1) -- (9,1);
    % Subdivide bottom-left quadrant
    \draw[thick] (7.5,0) -- (7.5,1);
    \draw[thick] (7,0.5) -- (8,0.5);
    % Mark high variance region
    \fill[pattern=north east lines, pattern color=oigray] (7,0) rectangle (7.5,0.5);
    \node[font=\tiny] at (7.25,0.25) {H};
    \node[below] at (8,0) {\small Level 3};

    % Legend
    \node[right, font=\small] at (9.2,1.5) {High $\sigma^2$:};
    \fill[pattern=north east lines, pattern color=oigray] (9.2,1) rectangle (9.6,1.3);
    \node[right, font=\small] at (9.7,1.15) {subdivide};
\end{tikzpicture}
\caption{Quadtree subdivision for hidden-node discovery. High-variance regions ($\sigma^2 > \tau$) are recursively subdivided; the hatched region (H) indicates where a hidden node may be discovered during pruning. Because the subdivision pattern depends on each CPPN's output field, different CPPNs terminate at different leaves, producing the per-CPPN variable-cardinality position sets at the root of the vectorization barrier.}
\label{fig:quadtree}
\end{figure}
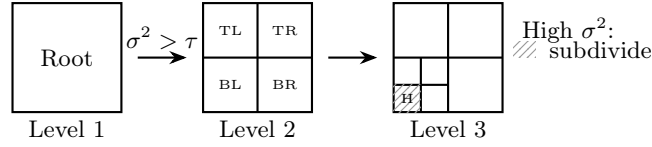

At each level, the procedure creates one child node at each of the four quadrant centers, queries the CPPN for their weights, and computes the variance. If variance exceeds $\tau_{\text{div}}$ (default 0.5), the quadrant subdivides further. Leaf nodes passing band-detection pruning become substrate hidden nodes. Figure~\ref{fig:quadtree} illustrates this variance-driven, data-dependent recursion.

\subsection{Pruning Extraction (Band Detection)}

The band threshold $\tau_{\text{band}}$ (default 0.3) controls connection extraction sensitivity. The formula $\max(\min(d_T, d_B), \min(d_L, d_R))$, where $d_T$, $d_B$, $d_L$, and $d_R$ are the absolute weight differences to the top, bottom, left, and right neighbors, identifies CPPN-output discontinuities: the inner $\min$ requires change along both directions of an axis (the node lies between weight regions), and the outer $\max$ accepts either horizontal or vertical boundaries. The feedforward constraint ensures connections flow from lower to higher y-coordinates, and Phase~2's hidden-to-hidden discovery respects it.

\subsection{Network Cleaning}

Network cleaning removes unreachable nodes via two-pass breadth-first search (BFS) reachability: forward from inputs along outgoing connections, backward from outputs along incoming ones. The intersection yields nodes on valid input-to-output paths; connections involving nodes outside it are discarded.

\section{Optimizations}
\label{sec:optimizations}

We develop four optimizations targeting the main bottlenecks in substrate discovery. Our unoptimized JAX-ESHN follows the three-phase discovery pipeline directly: the population-level \texttt{vmap}(transform) is included, but substrate discovery runs sequentially per CPPN with non-batched queries, standard BFS network cleaning, and sequential fitness evaluation. All speedups measure per-generation time relative to unoptimized JAX-ESHN on XOR at depth~2. Speedup values include interaction effects; individual contributions are not strictly additive.

\begin{itemize}
    \item \emph{O1 (Batched division queries).} Collects all children at each quadtree level and issues a single batched \texttt{vmap}(QueryCPPN) over a coordinate list of size $4 \times |Q_{\text{level}}|$, then variance-filters to decide which parents subdivide further.
    \item \emph{O2 (Batched pruning queries).} Collects neighbor coordinates across all leaves and issues a single mega-batch query: $4N$ coordinates reduce to one \texttt{vmap} call, with band-test results computed per leaf from the returned weights.
    \item \emph{O3 (vmap substrate evaluation).} Stacks all test inputs into a batch tensor and uses \texttt{vmap(ForwardPass)} for parallel evaluation, replacing sequential per-case computation.
    \item \emph{O4 (Precomputed coordinate offsets).} Precomputes the quadrant offset matrix $\mathbf{\Delta} \in \mathbb{R}^{4 \times 2}$ containing the four quadrant center offsets $(\pm 0.5, \pm 0.5)$ at initialization, allowing vectorized child position calculation via $\mathbf{pos} = (p_x, p_y) + \mathbf{\Delta} \times w$ using broadcasting.
\end{itemize}

Measured individually against the unoptimized implementation at population~1000, O1 provides 1.09$\times$, O2 1.23$\times$, and O3 1.30$\times$; O4, measured instead against the O2$+$O3 configuration, adds a further 1.01$\times$. The best combination we measured is O2$+$O3 at 1.73$\times$, and adding the remaining optimizations to it does not improve on that; a separate whole-implementation comparison at population~150 gives 1.65$\times$. Within-CPPN batching therefore plateaus near 1.7$\times$ however the optimizations are combined. Table~\ref{tab:optimizations} collects these measurements; because they come from two campaigns with different reference configurations, they are not a single cumulative ladder.

\begin{table}[t]
\centering
\caption{Per-generation speedup of the batched optimizations over unoptimized JAX-ESHN (XOR, depth~2). O1--O3 are measured individually against the unoptimized implementation at population~1000; O4 is measured against the O2$+$O3 configuration; the combination row is the best we measured. These come from separate campaigns with different reference configurations and are \emph{not} a cumulative ladder; they were recorded during development, and per-run files for them did not survive into the released archive. The ${\sim}1.7\times$ ceiling on within-CPPN batching is the empirical signature of the structural barrier: with population-level vectorization blocked, only within-CPPN work can be batched.}
\label{tab:optimizations}
\small
\begin{tabular*}{\textwidth}{@{\extracolsep{\fill}}l|c}
\toprule
\textbf{Optimization} & \textbf{Speedup} \\
\midrule
Unoptimized JAX-ESHN                       & 1.00$\times$ \\
O1 Batched division queries                & 1.09$\times$ \\
O2 Batched pruning queries                 & 1.23$\times$ \\
O3 \texttt{vmap} substrate evaluation      & 1.30$\times$ \\
O4 Precomputed coordinate offsets          & 1.01$\times$ \\
\midrule
O2 $+$ O3 (best measured combination)      & \textbf{1.73}$\times$ \\
\bottomrule
\end{tabular*}
\end{table}

Despite these gains, the per-CPPN variability described in Section~\ref{sec:introduction} still prevents population-level vectorization within static-shape compilation frameworks such as JAX/XLA, which keeps GPU utilization low for any implementation built on them. Section~\ref{sec:hshg} tests whether alternative data structures can escape this constraint; Section~\ref{sec:results} quantifies its impact.

\paragraph{Why not padding or static bounds?} Three approaches fail for \emph{distinct} reasons. \emph{Padding to maximum} yields uniform shapes but pads the \emph{output} of a traversal that still runs sequentially per CPPN, so the population loop survives intact; its memory cost ($O(N \times 4^d)$, over 99.9\% of it padding at depth 7 with population 500) is not what rules it out. Eagerly evaluating one shared grid pays the same asymptotic memory yet works, because it replaces the traversal instead of padding its result (Section~\ref{sec:emr}). \emph{Static position bounds} replace adaptive subdivision with uniform sampling, eliminating the variance-driven mechanism; Section~\ref{sec:hshg} documents this failure through HSHG, the prototypical instance. \emph{Dynamic recompilation} re-traces the XLA graph per shape change, defeating JIT caching. The padding and static-position routes both fail under fixed-size bounds, but by different mechanisms (a surviving sequential traversal vs.\ over-discovery); Section~\ref{sec:hshg} drills into the second because it shows \emph{why} adaptive discovery cannot be cheaply substituted.

\section{HSHG: An Attempted Solution to the Quadtree Bottleneck}
\label{sec:hshg}

Having established that per-CPPN discovery prevents vectorization (Section~\ref{sec:optimizations}), we tested replacing the quadtree with HSHG~\cite{teschner2003optimized}, a spatial-hashing structure with preallocated fixed-shape arrays. HSHG offers three properties critical for GPU acceleration: $O(1)$ average-case neighbor queries via hash-based cell lookup over a fixed 5$\times$5 candidate region (vs.\ $O(\log n)$ for quadtree), fixed-size state compatible with JAX JIT, and native \texttt{vmap} support through static memory allocation.

HSHG pre-generates all positions at all depth levels ($1+4+16+64=85$ positions at depth 3, per the formula in Section~\ref{sec:implementation}); a single mega-batch CPPN query retrieves all weights; per-position variance filtering then selects active nodes. This filter replaces both stages of the quadtree's adaptive discovery (variance-driven subdivision and band-detection pruning); HSHG retains neither. The core mismatch with how ES-HyperNEAT discovers nodes:
\begin{itemize}
    \item \emph{Quadtree}: IF variance $>$ threshold THEN subdivide $\rightarrow$ sparse, adaptive exploration
    \item \emph{HSHG}: Generate ALL positions $\rightarrow$ filter by variance $\rightarrow$ over-discovery
\end{itemize}

\subsection{Empirical Validation: Over-Discovery}

\begin{figure}[t]
\centering
\resizebox{0.9\textwidth}{!}{%
\begin{tikzpicture}[scale=0.6]
    % Title
    \node[font=\small\bfseries] at (6, 10.2) {Why HSHG Discovers Multiple Positions};
    % Vertical band showing spike width
    \fill[oiorange!10] (3.5, -0.7) rectangle (6.5, 9);
    % Top: CPPN with axes
    \draw[-{Stealth}, thick] (1, 7.3) -- (11, 7.3) node[right, font=\tiny] {$x$};
    \draw[-{Stealth}, thick] (1, 7.3) -- (1, 9.3) node[above, font=\tiny] {CPPN};
    \draw[thick, oigray] (1, 7.5) -- (3.5, 7.5) -- (5, 8.8) -- (6.5, 7.5) -- (11, 7.5);
    \fill[oiorange!30] (3.5, 7.3) -- (3.5, 7.5) -- (5, 8.8) -- (6.5, 7.5) -- (6.5, 7.3) -- cycle;
    \node[font=\tiny, oiorange, anchor=west] at (6.8, 8.6) {High variance};
    \draw[-{Stealth}, oiorange, thin] (6.7, 8.5) -- (5.3, 8.4);
    % Middle: HSHG fixed grid positions with neighborhoods
    \node[font=\scriptsize, oivermillion, anchor=west] at (0.3, 5.3) {HSHG:};
    \foreach \x in {2, 4, 6, 8, 10} {
        \draw[oigray!50] (\x, 5.8) -- (\x, 5.2);
    }
    \fill[oigray!50] (2, 5) circle (0.1);
    \fill[oivermillion] (4, 5) circle (0.12);
    \fill[oivermillion] (6, 5) circle (0.12);
    \fill[oigray!50] (8, 5) circle (0.1);
    \fill[oigray!50] (10, 5) circle (0.1);
    \node[font=\tiny, oigray] at (2, 4.6) {A};
    \node[font=\tiny, oivermillion] at (4, 4.6) {B};
    \node[font=\tiny, oivermillion] at (6, 4.6) {C};
    \node[font=\tiny, oigray] at (8, 4.6) {D};
    \node[font=\tiny, oigray] at (10, 4.6) {E};
    \fill[oivermillion!15] (3, 5.2) rectangle (5, 6.8);
    \draw[oivermillion, dashed] (3, 5.2) rectangle (5, 6.8);
    \node[font=\tiny, oivermillion, anchor=north west] at (3.05, 6.75) {B's zone};
    \fill[oiskyblue!20] (5, 5.2) rectangle (7, 6.8);
    \draw[oiblue, dashed] (5, 5.2) rectangle (7, 6.8);
    \node[font=\tiny, oiblue, anchor=north east] at (6.95, 6.75) {C's zone};
    \node[font=\tiny, oigray, align=left, anchor=west] at (-0.8, 6.4) {Fixed positions \\ sample CPPN \\ in their zones};
    \node[font=\tiny, oiorange, align=left, anchor=west] at (8, 6.0) {Both zones\\overlap spike!};
    \draw[-{Stealth}, oiorange, thin] (7.9, 6.2) -- (6.2, 6.2);
    \draw[-{Stealth}, oiorange, thin] (7.9, 5.8) -- (4.8, 5.8);
    \node[font=\tiny, oigray] at (2, 3.7) {low var};
    \node[font=\tiny, oigray] at (2, 3.3) {$\times$ skip};
    \node[font=\tiny, oivermillion] at (4, 3.7) {HIGH var};
    \node[font=\tiny, oivermillion] at (4, 3.3) {$\checkmark$ PASS};
    \node[font=\tiny, oivermillion] at (6, 3.7) {HIGH var};
    \node[font=\tiny, oivermillion] at (6, 3.3) {$\checkmark$ PASS};
    \node[font=\tiny, oigray] at (8, 3.7) {low var};
    \node[font=\tiny, oigray] at (8, 3.3) {$\times$ skip};
    \node[font=\tiny, oigray] at (10, 3.7) {low var};
    \node[font=\tiny, oigray] at (10, 3.3) {$\times$ skip};
    \node[font=\scriptsize, oivermillion, right] at (11, 5) {Found: \textbf{2}};
    % Bottom: quadtree comparison
    \node[font=\scriptsize, oiblue] at (0.8, 2) {Quadtree:};
    \draw[oiblue, thick] (2, 1.8) rectangle (10, 2.2);
    \node[font=\tiny, oiblue] at (6, 2) {whole space};
    \draw[oiblue, thick] (2, 1) rectangle (6, 1.4);
    \draw[oigray!30] (6, 1) rectangle (10, 1.4);
    \node[font=\tiny, oiblue] at (4, 1.2) {left half};
    \node[font=\tiny, oigray!50] at (8, 1.2) {ignore};
    \draw[oiblue, thick] (3.5, 0.2) rectangle (6, 0.6);
    \node[font=\tiny, oiblue] at (4.75, 0.4) {subdivide};
    \fill[oiblue] (5, -0.2) circle (0.15);
    \node[font=\tiny, oiblue] at (5, -0.55) {creates position HERE};
    \node[font=\tiny, oiorange, align=center, anchor=west] at (7, -0.2) {Converges to\\this position};
    \draw[-{Stealth}, oiorange, thin] (6.9, -0.2) -- (5.3, -0.2);
    \node[font=\scriptsize, oiblue, right] at (11, 1) {Found: \textbf{1}};
\end{tikzpicture}}
\caption{Why HSHG over-discovers positions. \textbf{Top:} a CPPN with one informative feature (spike). \textbf{Middle:} HSHG pre-generates fixed positions (A--E), each checking CPPN variance \emph{independently} in its own zone; positions B and C both detect high variance because their zones overlap the spike, so both pass and discover two positions for one feature. \textbf{Bottom:} the quadtree subdivides \emph{toward} variance, creating exactly one position at the feature. The contrast is independent per-position filtering (HSHG) versus hierarchical, parent-gated filtering (quadtree); the latter is what EMR-HyperNEAT preserves (Section~\ref{sec:emr}).}
\label{fig:mechanism}
\end{figure}
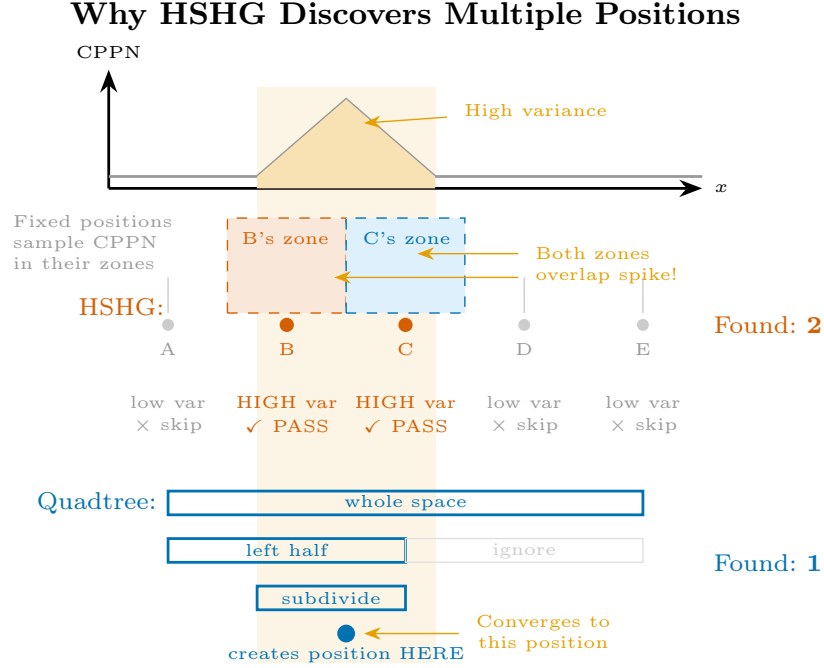

We tested HSHG on the XOR benchmark. Because pre-generated positions sample fixed neighborhoods, adjacent positions can overlap the same CPPN feature: when two HSHG positions' sampling neighborhoods both intersect a single informative spike, both positions independently detect high variance, yielding two hidden nodes for one feature; the quadtree by contrast would adaptively subdivide and create just one. Figure~\ref{fig:mechanism} makes the mechanism explicit.

We tested multiple configurations (variance thresholds 0.03--0.5, grid depths 2--4) across 50 CPPN evaluations at the substrate-discovery level. This probe measures discovery directly, under a different protocol from the 30-seed evolutionary campaign of Table~\ref{tab:hshg_task_performance}, and its per-CPPN records are not part of the released campaign archive. HSHG over-discovers by an order of magnitude: 29--63 hidden nodes against the quadtree's 1--2. Fitness does not recover: best fitness 0.746 came from a CPPN that discovered only 1 node, while CPPNs discovering 62 nodes plateaued at 0.500. No parameter adjustment recovered quadtree-level sparsity. The order-of-magnitude over-discovery is sufficient to diagnose HSHG's failure: networks with 62 hidden nodes for XOR are pathologically bloated regardless of fitness outcome. The quadtree's variance-driven subdivision, which examines only where variance indicates information, is what makes ES-HyperNEAT's sparse topology discovery work. The threshold-mismatch between the two systems quantifies this: the Baseline's default quadtree division threshold (0.5) controls \emph{whether to explore further}, while the variance threshold (0.03), which HSHG repurposes as its per-position activity test, controls \emph{whether a pre-explored position is active}; the $16.7\times$ sensitivity gap reflects this role inversion, not a tuning error.

\section{Experimental Results}
\label{sec:results}

We benchmark JAX-ESHN against the CPU Baseline on XOR, comparing solve rates, generation times, and \emph{construction overhead}, the cost of the first evolutionary step (Section~\ref{subsec:construction-overhead} defines and decomposes it). The main finding is that construction overhead dominates runtime at shallow depths, while JAX-ESHN wins at deeper substrates where the Baseline becomes prohibitively slow.

\subsection{Benchmark Configuration}

We evaluate on XOR, a standard ES-HyperNEAT benchmark~\cite{risi2012enhanced} whose
non-linearly-separable decision boundary requires at least one adaptively discovered hidden layer (a
nonlinear hidden representation), which ES-HyperNEAT supplies through quadtree subdivision. Both
implementations run at \texttt{iteration\_level}~1 and both solve it, so the comparison is not
confounded by a difference in expansion depth. Both implementations test maximum depths 1--7 (D1--D7, spanning 5 to 21,845 positions) at initial depth 0, with 3 seeds per configuration. The two campaigns used different fitness thresholds and disjoint seed sets (Baseline 0.99 with seeds 42/123/456, JAX-ESHN 0.98 with 42/43/44), which Appendix~\ref{sec:repro} tabulates and quantifies. The Baseline runs on an Apple~M4~Max (CPU) with 8 population sizes (50--1000); JAX-ESHN on NVIDIA RTX 2080 Ti with 10 population sizes (50--1000). Solve rates (Table~\ref{tab:runtime_comparison}) aggregate 8 population sizes per implementation, from 300-generation runs for JAX-ESHN and 30-generation runs for the Baseline. JAX-ESHN timing (Tables~\ref{tab:runtime_comparison}--\ref{tab:jit_by_pop}) uses 10-generation forced reruns (fresh timing runs) across all 10 population sizes, from which the 30-generation totals are projected (Appendix~\ref{sec:repro}). JAX-ESHN uses all optimizations from Section~\ref{sec:optimizations}.

\subsection{Statistical Methods}

Proportion differences (solve rates) are compared using Fisher's exact test. Continuous outcomes (generation times) are analyzed with one-way ANOVA; we report $F$ and effect size~$\eta^2$. Discrete outcomes (depth effects on solve rate) use a $\chi^2$ test of independence. Correlations between continuous variables are Pearson $r$. The polynomial regression of generation time on depth and population is summarized by $R^2$ with 5-fold cross-validation. All findings are significant at $p < 0.05$ unless stated otherwise; exact $p$-values are reported at first occurrence. Because the principal effects carry $p$-values far below this threshold (depth-on-time $F$-test $p<10^{-71}$, Parity-3 Fisher $p<10^{-5}$, depth-on-solve-rate $\chi^2$ $p\approx1.3\times10^{-7}$), no Bonferroni or Holm correction over the handful of tests reported here alters any conclusion, so we report uncorrected exact $p$-values; the weaker population--solve-rate correlation ($r=0.27$) is reported as a trend, not a corrected effect.

\subsection{Runtime and Solve Rate Comparison}

Table~\ref{tab:runtime_comparison} compares total runtime at Pop~500 with solve rates aggregated across eight population sizes, for depths D3--D7.

\begin{table}[h]
\centering
\caption{XOR runtime (minutes) and solve rate at Pop~500, depths D3--D7. The JAX-ESHN \emph{Total} is projected over 30 generations as construction overhead plus 29 further generations at the measured post-construction per-generation rate, since generation~1 is already inside construction overhead ($n=3$ forced reruns; see Appendix~\ref{sec:repro}). \emph{Solve} is the analogous projection to the solve generation rather than to generation~30, so a run that solves in generation~1 costs exactly its construction overhead. Solve rates aggregated across 8 population sizes (168 trials per implementation across D1--D7, 24 per depth, of which the 120 for D3--D7 are shown; 300-generation budget for JAX-ESHN, 30-generation for the Baseline).}
\label{tab:runtime_comparison}
\small
\begin{tabular*}{\textwidth}{@{\extracolsep{\fill}}r|rc|rrc}
\toprule
& \multicolumn{2}{c|}{\textbf{Baseline}} & \multicolumn{3}{c}{\textbf{JAX-ESHN}} \\
\textbf{D} & \textbf{Time} & \textbf{Solve\%} & \textbf{Total} & \textbf{Solve} & \textbf{Solve\%} \\
\midrule
3 & $0.3 \pm 0.1$   & 62.5  & $180 \pm 6$       & $12 \pm 3$    & 100 \\
4 & $1.2 \pm 0.6$   & 37.5  & $434 \pm 47$      & $39 \pm 14$   & 91.7 \\
5 & $11 \pm 4$      & 33.3  & $1{,}618 \pm 462$ & $128 \pm 17$  & 100 \\
6 & $110 \pm 80$    & 8.3   & $2{,}500 \pm 521$ & $215 \pm 58$  & 100 \\
7 & $824 \pm 528$   & 4.2   & $3{,}540 \pm 562$ & $348 \pm 157$ & 100 \\
\bottomrule
\end{tabular*}
\end{table}

\begin{figure}[!b]
\centering
\begin{tikzpicture}
\begin{axis}[
    width=0.99\textwidth,
    height=6cm,
    ylabel={Total Runtime (30 gen)},
    xlabel={Depth},
    ymode=log,
    ytick={1, 10, 60, 600, 3600, 36000, 360000},
    yticklabels={1s, 10s, 1min, 10min, 1h, 10h, 100h},
    ymin=0.5,
    ymax=500000,
    xmin=0.5,
    xmax=10.5,
    xtick={1,2,3,4,5,6,7,8,9,10},
    ymajorgrids=true,
    grid style={dashed, gray!30},
    tick label style={font=\small},
    label style={font=\small},
    legend style={at={(0.98,0.02)}, anchor=south east, font=\tiny},
]
% Baseline IQR band (25th-75th percentile across 8 population sizes)
% Uniform D1-D7: iteration_level==1, per-population mean of total_time_s,
% then 25th/75th percentile across the 8 population means. D6 uses the 18
% iteration-level-1 runs in pureples_depth6_partial.md (total = per-gen x 30).
% D7 follows the same rule as D1-D6.
\addplot[name path=baselinemin, forget plot, color=oivermillion, thick, dotted] coordinates {
    (1, 1.9) (2, 2.5) (3, 8.6) (4, 49.6) (5, 191.5) (6, 2716.6) (7, 15854.7)
};
\addplot[name path=baselinemax, forget plot, color=oivermillion, thick, dotted] coordinates {
    (1, 6.0) (2, 7.5) (3, 20.2) (4, 81.1) (5, 492.8) (6, 6100.4) (7, 41385.8)
};
\addplot[fill=oivermillion, fill opacity=0.2] fill between[of=baselinemin and baselinemax];

% JAX-ESHN IQR band (25th-75th percentile across 10 populations, 30-gen total)
% Convention: construction overhead + 29 x post-construction per-gen, matching
% tab:runtime_comparison and the Data-provenance paragraph. Generation 1 is inside
% construction overhead (total_evolution/per_gen == gens-1), so a 30-generation
% total is construction + 29 further generations.
\addplot[name path=jaxmin, forget plot, color=oiblue, thick, dotted] coordinates {
    (1, 307.2) (2, 1044.9) (3, 3602.9) (4, 10377.2) (5, 44035.3) (6, 59270.6) (7, 82481.7)
};
\addplot[name path=jaxmax, forget plot, color=oiblue, thick, dotted] coordinates {
    (1, 981.1) (2, 3315.7) (3, 12226.3) (4, 36212.3) (5, 117699.0) (6, 209456.7) (7, 251050.4)
};
\addplot[fill=oiblue, fill opacity=0.2] fill between[of=jaxmin and jaxmax];

% JIT compilation time IQR band (25th-75th percentile across 10 population sizes)
\addplot[name path=jitmin, forget plot, color=oigreen, thick, dotted] coordinates {
    (1, 25.9) (2, 70.5) (3, 149.0) (4, 436.0) (5, 1327.0) (6, 2004.3) (7, 2099.5)
};
\addplot[name path=jitmax, forget plot, color=oigreen, thick, dotted] coordinates {
    (1, 82.1) (2, 235.9) (3, 649.5) (4, 1767.3) (5, 5110.5) (6, 6024.0) (7, 6925.5)
};
\addplot[fill=oigreen, fill opacity=0.15] fill between[of=jitmin and jitmax];

% JIT compilation time - Pop 1000 (solid line, circle markers)
% D1-D7: measured. D8-D10: the separate 300-gen campaign, so the line is
% broken between D7 and D8.
\addplot[color=oigreen, thick, mark=*, solid] coordinates {
    (1, 86) (2, 321) (3, 975) (4, 3110) (5, 9986) (6, 13700) (7, 16151)
};
\addplot[color=oigreen, thick, mark=*, solid, forget plot] coordinates {
    (8, 14034) (9, 16575) (10, 21955)
};

% Baseline Time to Solve - Pop 1000 (solid line, triangle markers)
% D6 omitted: no seeds solved (30.00 avg generations = timeout)
\addplot[color=oivermillion, thick, mark=triangle*, solid] coordinates {
    (1, 4.1) (2, 5.2) (3, 13.1) (4, 66.9) (5, 1279.8) (7, 55206.6)
};

% JAX-ESHN Time to Solve - Pop 1000 (solid line, hollow square markers)
% Cost to solve at generation k = construction + (k-1) x per-gen, because generation 1
% is inside construction overhead. A gen-1 solve therefore costs exactly construction,
% which is why D2/D4/D6/D7 coincide with the construction line (all 3 seeds solve at
% gen 1); D3/D5 sit higher because one seed each solves at generation 2.
% D1: the 10-gen campaign never solves it, so D1 uses the 300-gen campaign's mean solve
% generation (13.33) at this campaign's per-generation rate = 941.1 s (measured 958.5 s).
% D8-D10: solves at gen 1 in the separate 300-gen campaign, so solve time =
% JIT time; the line is broken between D7 and D8.
\addplot[color=oiblue, thick, mark=square, solid] coordinates {
    (1, 941.1) (2, 320.7) (3, 1180.4) (4, 3110.4) (5, 12409.3) (6, 13700.3) (7, 16151.3)
};
\addplot[color=oiblue, thick, mark=square, solid, forget plot] coordinates {
    (8, 14034) (9, 16575) (10, 21955)
};

% Vertical line separating multi-pop data (D1-D7) from Pop-1000-only (D8-D10)
\draw[dashed, gray, thick] (axis cs:7.5,0.5) -- (axis cs:7.5,500000);

\legend{Baseline IQR, JAX-ESHN IQR, Constr. IQR, Constr. (Pop 1000), Baseline Solve (Pop 1000), JAX-ESHN Solve (Pop 1000)}
\end{axis}
\end{tikzpicture}
\caption{XOR benchmark: total runtime comparison (log scale).
\textbf{Shaded bands} (D1--D7): interquartile range (IQR) for Baseline CPU (orange), JAX-ESHN 30-gen total (blue), construction overhead only (green).
\textbf{Solid lines} (Pop 1000, D1--D10): construction overhead (circles), Baseline solve time (triangles), JAX-ESHN solve time (squares).
Dashed line separates D1--D7 multi-population data from D8--D10 Pop-1000-only (separate measurement campaigns). Baseline D6 is omitted from the solve line (no Pop-1000 seed solved); every remaining Baseline point averages in a seed that hit the 30-generation cap without solving (2 of 3 seeds solved at D1--D5, 1 of 3 at D7), so all of them are lower bounds on the Baseline's true solve time. JAX-ESHN's D1 solve point combines the 300-generation campaign's mean solve generation (${\sim}13$) with this campaign's per-generation rate; at D2--D7, 16 of 18 seeds solve during the construction generation itself, so those points sit at or barely above the construction line, which is why the solve curve dips from D1 to D2.}
\label{fig:baseline_iqr}
\end{figure}
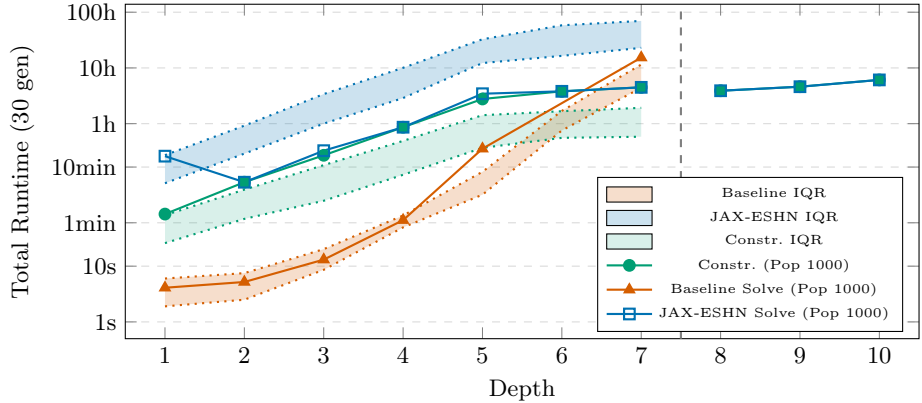

JAX-ESHN achieves near-100\% solve rates across D3--D7 (only D4 at 91.7\%, likely statistical variation at $n=24$), while the Baseline peaks at 62.5\% (D3) and drops to 4.2\% (D7); the Baseline requires extended evolution and cannot complete it economically at D6+ (nearly 14 hours for 30 generations at D7). The gen-1 solve phenomenon at deeper substrates is detailed in Section~\ref{subsec:construction-cost} (Figure~\ref{fig:baseline_iqr}). The 300-versus-30-generation budget asymmetry does not drive this gap: under a 30-generation cap, JAX-ESHN's aggregated solve rates would still be $95.8/79.2/95.8/100/95.8\%$ across D3--D7 (from the campaign's per-run generation counts), above the Baseline at every depth. The symmetric question cannot be answered as cleanly: the Baseline's own rates are budget-sensitive, and at 100 generations it solves XOR in 30 of 30 seeds at D2--D3 (Table~\ref{tab:hshg_task_performance}), so its D6--D7 rates here should be read as rates \emph{under a 30-generation cap} rather than as a ceiling. Extending that budget at D6--D7 was not affordable at ${\approx}14$ hours per run.

At D3--D5, the Baseline is roughly $150$--$600\times$ faster in wall-clock time, but its solve rate drops steadily. The solve-time column tells a different story: JAX-ESHN solves in 12--128 minutes (construction plus 0--2 further generations) while the Baseline must run many generations through its expensive quadtree traversal, and at D6--D7 usually fails anyway. At D7, the Baseline takes 824 minutes and solves at 4.2\%; JAX-ESHN solves in $348 \pm 157$ minutes at 100\%. Baseline runtime scales ${\approx}7\times$ per depth level. The solve-rate fragility is an inflection, not a smooth slope: the steepest drop sits between D5 (33.3\%) and D6 (8.3\%), corresponding to the $4\times$ jump in candidate positions (1{,}365 to 5{,}461) that pushes most seeds past the 30-generation timeout. For practical guidance, see Section~\ref{sec:post_jit}.

Because the deep-substrate timing rests on only $n=3$ seeds, these point estimates carry wide intervals. Every dispersion we report is a population standard deviation, while the $t$-intervals below use the sample standard deviation, so the two differ by a factor of $\sqrt{3/2}$ at $n=3$: the projected 30-generation total at D5 (Pop~500) is $1{,}618$ minutes with a 95\% $t$-interval of $[213,\,3{,}022]$, widening to $[1{,}830,\,5{,}251]$ at D7. We therefore read D5--D7 \emph{timing} as indicative of magnitude and direction, not precise estimates, and base distributional claims on the $n=30$ multi-benchmark campaign (Figure~\ref{fig:perf_distribution}), where per-seed variability is characterized directly.

\paragraph{Statistical validation (Baseline).} Statistical analysis of 456 Baseline XOR benchmark trials (D1--D6: 6 depths $\times$ 8 populations $\times$ 3 iteration levels $\times$ 3 replications; D7 limited to iteration level 1 due to computational constraints) confirms that depth significantly affects both generation time ($F = 87.04$, $\eta^2 = 0.54$, $p < 10^{-71}$) and solve rate ($\chi^2 = 42.73$, $p \approx 1.3\times10^{-7}$), while population has a moderate positive effect on solve rate ($r = 0.27$). The position-to-time correlation ($r = 0.73$) directly supports the quadtree bottleneck hypothesis. High variance in total runtime reflects different numbers of generations required to solve XOR across replications.

A polynomial regression model predicting generation time from depth and population achieves $R^2 = 0.95$ (cross-validation: $0.85 \pm 0.11$), confirming the deterministic nature of computational overhead.

\subsection{Construction Overhead: Compilation Plus First Generation (JAX-ESHN)}
\label{subsec:construction-overhead}

We report \emph{construction overhead}: the wall-clock duration of the first evolutionary step, which comprises XLA compilation and the first-generation evaluation that runs immediately afterward. We adopt this name rather than ``JIT time'' because the quantity is not pure compilation; the two components are bundled in JAX's lazy-compilation timing, and we examine their separation below. Figure~\ref{fig:baseline_iqr} reports it for all 10 population sizes (50--1000) at D1--D7; Table~\ref{tab:jit_by_pop} tabulates five of them at D3--D7.

Construction overhead grows exponentially with depth and approximately linearly with population size at fixed depth. At shallow depths (D1--D2), it remains under 6 minutes (Figure~\ref{fig:baseline_iqr}); at depth 3, it reaches 16 minutes for Pop 1000 (Table~\ref{tab:jit_by_pop}), but at depth 6, the initial generation takes 3+ hours before evolution can proceed. The overhead increases roughly $3\times$ per depth level through depth 5; the per-level growth then slows sharply ($1.37\times$ at D5$\to$D6 and $1.18\times$ at D6$\to$D7) and remains within the plateau through the separate 300-generation D8--D10 campaign ($1.18\times$ at D8$\to$D9 and $1.32\times$ at D9$\to$D10). Throughout this plateau the cost trajectory falls well below the $\sim$$4^d$ growth in quadtree position count, easing rather than worsening JAX-ESHN's dominant cost driver.
\paragraph{Exploratory deep substrates (D8--D10).}\label{subsec:exploratory_deep} Under computational constraints, depths 8--10 were tested only at Pop 1000 with $n=3$ replications (exploratory; limited sample size). Depth 8 (100\% solve) shows construction overhead $14{,}034 \pm 633$s ($\sim$3.9h); depth 9 (100\% solve) shows $16{,}575 \pm 387$s ($\sim$4.6h, only 18\% slower than D8 despite 4$\times$ more positions); depth 10 (100\% solve) shows $21{,}955 \pm 272$s ($\sim$6.1h, 32\% slower than D9). All three remain within the plateau rather than resuming pre-plateau growth.

\paragraph{Construction, not search, is the cost.}\label{subsec:construction-cost} Across the full experimental grid, JAX-ESHN's cost is dominated by substrate construction (JIT + first generation). At Pop~$\geq 750$, JAX-ESHN solves XOR during the construction generation itself at depths D4--D7, and Pop~1000 extends this pattern through D10. The same pattern inverts the depth-difficulty intuition: at Pop~50, mean generations-to-solve falls from $155.3$ at D1 (one unsolved seed counted at the 300-generation cap) to $6.0$ at D7, the opposite of the Baseline's depth-decay direction (Table~\ref{tab:runtime_comparison}). Deeper substrates carry more representational headroom, and once that headroom exists, search is essentially trivial. Read together, the two observations reframe the bottleneck: JAX-ESHN's cost scales with depth because constructing those substrates is expensive, not because evolution needs deeper substrates to solve XOR.

\paragraph{Population scaling and component isolation.}\label{subsec:isolation} Across D3--D7, Pop~1000 construction overhead is 18--23$\times$ larger than at Pop~50, closely matching the 20$\times$ population ratio. Standard lazy-compilation timing bundles construction overhead's two components, but they \emph{can} be separated: JAX's ahead-of-time interface (\texttt{jax.jit(f).lower(*args).compile()}) times pure compilation, and the first call to the compiled function times execution. The population-linearity has a clear structural source: each CPPN's unique substrate topology requires its own compilation and evaluation, so first-generation work iterates sequentially over the $N$ CPPNs and contributes a term linear in population by construction. This points to first-generation work, rather than a super-linear tracing cost, as the bulk of the population multiplier.

Depth growth acts through a different route than population. We test this directly with an isolation harness built on that ahead-of-time interface, timing pure XLA compilation separately from execution (its output and the summarizing script are released; Appendix~\ref{sec:repro}). At depth~1 ($n=3$ seeds per population), with population swept 20-fold (50--1000), pure compilation of the population-sized NEAT-update graph stays flat (a $1.07\times$ ratio), while first-generation evaluation, a sequential per-CPPN Python loop, scales $16.9\times$ (near-linear). The population multiplier in construction overhead is therefore the evaluation loop. Depth instead enlarges each \emph{individual's} workload: a deeper substrate subdivides the quadtree further (more CPPN queries per individual) and yields larger discovered tensors (inflating the per-CPPN forward-pass compile on GPU), whereas the population-sized NEAT-update compile depends on CPPN-genome shape and is roughly flat in depth as well. The quadtree subdivision itself is host-side Python recursion, so what compiles are the per-CPPN forward passes and the NEAT \texttt{ask}/\texttt{tell} steps. We ran these probes on the CPU backend under heavy concurrent host load, so we report the population and depth \emph{ratios}, which are insensitive to a roughly constant load factor, rather than absolute timings.

\begin{table}[h]
\centering
\caption{JAX-ESHN construction overhead (compilation $+$ first-generation evaluation; minutes, mean $\pm$ std) by depth and population ($n=3$ seeds).}
\label{tab:jit_by_pop}
\small
\begin{tabular*}{\textwidth}{@{\extracolsep{\fill}}r|rrrrr}
\toprule
\textbf{D} & \textbf{Pop 50} & \textbf{Pop 150} & \textbf{Pop 300} & \textbf{Pop 500} & \textbf{Pop 1000} \\
\midrule
3 & $0.9 \pm 0.0$  & $2.5 \pm 0.1$  & $4.4 \pm 0.1$   & $7.7 \pm 0.1$   & $16.3 \pm 1.0$ \\
4 & $2.5 \pm 0.2$  & $7.3 \pm 0.3$  & $12.4 \pm 0.5$  & $23.7 \pm 0.7$  & $51.8 \pm 1.8$ \\
5 & $7.3 \pm 1.3$  & $22.1 \pm 1.5$ & $36.7 \pm 1.4$  & $74.4 \pm 0.8$  & $166.4 \pm 10.5$ \\
6 & $11.8 \pm 1.8$ & $33.4 \pm 1.8$ & $52.1 \pm 0.5$  & $100.4 \pm 0.3$ & $228.3 \pm 5.9$ \\
7 & $12.3 \pm 1.8$ & $35.0 \pm 1.8$ & $55.5 \pm 0.6$  & $105.9 \pm 0.4$ & $269.2 \pm 30.3$ \\
\bottomrule
\end{tabular*}
\end{table}

\subsection{Multi-Benchmark Validation}
\label{sec:multi_benchmark}

To test whether the quadtree bottleneck holds across task types, we ran both implementations on four further benchmarks spanning Boolean, continuous, and control problems: Parity-3 (Boolean, 3 inputs), circle classification (continuous, 2 inputs), sine regression (continuous, 1 input), and CartPole (reinforcement-learning control, 4 inputs). The three cheap tasks were run at depths 2--4 with 30 seeds per condition (540 runs: 270 Baseline, 270 JAX-ESHN; Tables~\ref{tab:multi_benchmark}--\ref{tab:multi_benchmark_jax}); CartPole (180 further runs) is analyzed separately in Section~\ref{sec:cartpole} because its heavy fixed evaluation cost and the two implementations' differing stopping behavior call for a per-generation comparison rather than a total-runtime one.

\paragraph{A CPU-vs-CPU control.} Unlike the XOR scaling study of Section~\ref{sec:results} (where JAX-ESHN ran on the GPU), the JAX-ESHN runs here used JAX's \emph{CPU} backend, while the Baseline used neat-python on CPU. Both implementations therefore ran on CPU, so any depth-scaling divergence observed below is a property of the substrate-discovery \emph{implementation}, not of GPU acceleration. This directly addresses the concern that the GPU-vs-CPU hardware difference in the XOR study could confound the structural conclusion; the only residual difference here is the CPU model (Apple Silicon vs.\ x86, on separate machines), which Section~\ref{sec:caveats} discusses.

\begin{table}[h]
\centering
\caption{Baseline multi-benchmark results (Pop 150, $n=30$, 100 gen). Solve rate (\%) and mean $\pm$ std time per seed (D2--D3 in seconds, D4 in hours). The d2$\to$d4 column is the D4/D2 mean-time ratio.}
\label{tab:multi_benchmark}
\small
\begin{tabular*}{\textwidth}{@{\extracolsep{\fill}}l|rr|rr|rr|r}
\toprule
& \multicolumn{2}{c|}{\textbf{Depth 2}} & \multicolumn{2}{c|}{\textbf{Depth 3}} & \multicolumn{2}{c|}{\textbf{Depth 4}} & \\
\textbf{Task} & \textbf{\%} & \textbf{Time} & \textbf{\%} & \textbf{Time} & \textbf{\%} & \textbf{Time} & \textbf{d2$\to$d4} \\
\midrule
Parity-3   & 76.7 & $194 \pm 126$    & 100  & $953 \pm 741$      & 96.7 & $3.8 \pm 4.8$\,h  & 71$\times$ \\
Circle     & 0    & $291 \pm 60$     & 0    & $2{,}225 \pm 804$  & 0    & $5.2 \pm 5.5$\,h  & 65$\times$ \\
Sine       & 0    & $245 \pm 57$     & 0    & $1{,}971 \pm 626$  & 0    & $4.5 \pm 5.9$\,h  & 67$\times$ \\
\bottomrule
\end{tabular*}
\end{table}

\begin{table}[h]
\centering
\caption{JAX-ESHN multi-benchmark results (JAX \emph{CPU} backend; see Section~\ref{sec:multi_benchmark}). Same config as Table~\ref{tab:multi_benchmark}. Times (mean $\pm$ std) include construction overhead (compilation $+$ first generation); D2--D3 in minutes, D4 in hours. The d2$\to$d4 column is the D4/D2 mean-time ratio.}
\label{tab:multi_benchmark_jax}
\small
\begin{tabular*}{\textwidth}{@{\extracolsep{\fill}}l|rr|rr|rr|r}
\toprule
& \multicolumn{2}{c|}{\textbf{Depth 2}} & \multicolumn{2}{c|}{\textbf{Depth 3}} & \multicolumn{2}{c|}{\textbf{Depth 4}} & \\
\textbf{Task} & \textbf{\%} & \textbf{Time} & \textbf{\%} & \textbf{Time} & \textbf{\%} & \textbf{Time} & \textbf{d2$\to$d4} \\
\midrule
Parity-3    & 13.3 & $23 \pm 2$\,m   & 23.3 & $47 \pm 6$\,m   & 16.7 & $1.9 \pm 0.4$\,h  & 4.9$\times$ \\
Circle      & 0    & $28 \pm 2$\,m   & 0    & $57 \pm 12$\,m  & 0    & $2.3 \pm 0.7$\,h  & 4.8$\times$ \\
Sine        & 96.7 & $23 \pm 2$\,m   & 100  & $47 \pm 7$\,m   & 100  & $1.9 \pm 0.5$\,h  & 5.1$\times$ \\
\bottomrule
\end{tabular*}
\end{table}

\begin{figure}[t]
\centering
\begin{tikzpicture}
\begin{axis}[
    boxplot/draw direction=y,
    boxplot/box extend=0.5,
    ymode=log,
    width=\textwidth,
    height=5.6cm,
    ylabel={Per-generation cost (s, log)},
    ymin=25, ymax=2000,
    xmin=0.3, xmax=6.4,
    xtick={1.35, 3.35, 5.35},
    xticklabels={Parity-3, Circle, Sine},
    tick label style={font=\small},
    label style={font=\small},
    ymajorgrids=true,
    grid style={dashed, gray!30},
    legend style={at={(0.02,0.97)}, anchor=north west, font=\tiny},
]
% Baseline (CPU): oivermillion
\addplot+[boxplot prepared={draw position=1, lower whisker=45.2, lower quartile=102.4, median=179.1, upper quartile=707.6, upper whisker=1155}, oivermillion, solid, fill=oivermillion!20, area legend] coordinates {};
\addlegendentry{Baseline (CPU)}
\addplot+[boxplot prepared={draw position=3, lower whisker=51.0, lower quartile=83.5, median=99.4, upper quartile=161.4, upper whisker=792.6}, oivermillion, solid, fill=oivermillion!20, forget plot] coordinates {};
\addplot+[boxplot prepared={draw position=5, lower whisker=33.4, lower quartile=63.4, median=73.9, upper quartile=123.3, upper whisker=968.3}, oivermillion, solid, fill=oivermillion!20, forget plot] coordinates {};
% JAX-ESHN (CPU backend): oiblue
\addplot+[boxplot prepared={draw position=1.7, lower whisker=43.0, lower quartile=53.6, median=64.5, upper quartile=78.2, upper whisker=93.8}, oiblue, solid, fill=oiblue!20, area legend] coordinates {};
\addlegendentry{JAX-ESHN (CPU)}
\addplot+[boxplot prepared={draw position=3.7, lower whisker=45.4, lower quartile=63.8, median=74.8, upper quartile=87.1, upper whisker=163.8}, oiblue, solid, fill=oiblue!20, forget plot] coordinates {};
\addplot+[boxplot prepared={draw position=5.7, lower whisker=40.8, lower quartile=53.6, median=59.5, upper quartile=74.8, upper whisker=105.9}, oiblue, solid, fill=oiblue!20, forget plot] coordinates {};
\end{axis}
\end{tikzpicture}
\caption{Per-generation cost distributions at depth~4 (Pop~150, $n=30$ seeds), both implementations on CPU. Per-generation cost is measured per implementation as each one's steady-state rate: total runtime over generations executed for the Baseline, and the post-construction rate for JAX-ESHN, so JAX-ESHN's one-time construction overhead is excluded here. Amortizing it in would raise JAX-ESHN's medians by only 4--6\% (to 68.1, 77.8, and 62.2~s), leaving the contrast unchanged. Boxes show the interquartile range and median; whiskers span the full min--max range. JAX-ESHN's distributions are tight (coefficient of variation $0.22$--$0.31$), while the Baseline's are right-skewed and bimodal (CV $0.90$--$1.30$): most seeds finish quickly, but a long upper tail grinds through many expensive quadtree generations. This predictability gap is the distributional counterpart to the scaling contrast, and uses the same per-generation metric (Section~\ref{sec:cartpole}) to neutralize the implementations' differing stopping behavior.}
\label{fig:perf_distribution}
\end{figure}
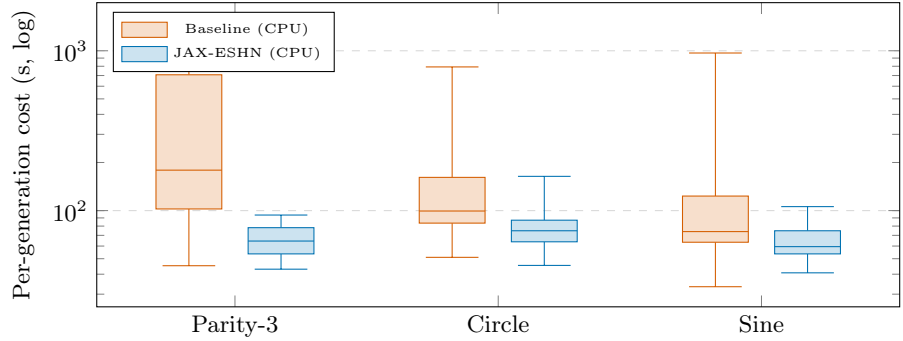

\paragraph{Scaling contrast.} Tables~\ref{tab:multi_benchmark}--\ref{tab:multi_benchmark_jax} show the incompatibility from opposite directions: the Baseline scales 65--71$\times$ from D2 to D4 because quadtree traversal dominates, while JAX-ESHN, whose within-CPPN queries are batched (Section~\ref{sec:optimizations}), scales only 4.8--5.1$\times$. Because both implementations ran on CPU in this campaign, this divergence isolates the implementation from the hardware: the quadtree's per-generation cost, not the choice of GPU versus CPU, is what scales steeply. Task type, input dimensionality, and fitness function have no measurable effect on either scaling exponent. JAX-ESHN is also far more predictable: at D4, its coefficient of variation (std/mean of total runtime) is 0.2--0.3 across tasks, while the Baseline reaches 1.0--1.3 because early-solving and timeout seeds produce a bimodal runtime distribution. Steady-state per-generation comparison at this same operating point sharpens the picture from a different angle than the D7 reliability advantage discussed in Section~\ref{sec:post_jit}. At D4 (Pop~150), once construction overhead amortizes, JAX-ESHN's mean per-generation cost is $2.4\times$ lower than the Baseline's on circle classification, $2.5\times$ on sine regression, and $6.1\times$ on Parity-3 (computed by isolating post-construction per-generation timing from the same multi-benchmark runs). These are ratios of means; because the Baseline's distribution is right-skewed, the corresponding ratios of the medians plotted in Figure~\ref{fig:perf_distribution} are smaller ($1.3\times$, $1.2\times$, and $2.8\times$). The spread tracks how heavy each task's fitness evaluation is. The post-JIT advantage exists but is task-dependent and far short of the order-of-magnitude gain population-level vectorization would deliver.

\paragraph{Solve-rate differences.} The implementations disagree sharply on which tasks are solvable. The Baseline solves Parity-3 at 77--100\% while JAX-ESHN manages only 13--23\% ($p < 10^{-5}$). Conversely, JAX-ESHN solves sine regression at 97--100\% while the Baseline never approaches the threshold, peaking at 0.752. Neither solves circle classification at any depth: both plateau just below threshold at essentially the same fitness (Baseline max $0.89$--$0.90$, JAX-ESHN max $0.887$), and this \emph{agreement} on failure, unlike the library-driven disagreement on Parity-3 and sine, points to a representational ceiling both implementations share, not a configuration artifact. It is orthogonal to the scaling claim, which holds on circle regardless of solve rate (Tables~\ref{tab:multi_benchmark}--\ref{tab:multi_benchmark_jax}). By contrast, both implementations solve CartPole at 100\% across all depths (Section~\ref{sec:cartpole}), the one task on which they fully agree; this is a useful positive control, because it shows that JAX-ESHN's low Parity-3 solve rate reflects a library/task interaction rather than a capacity ceiling of the discovered substrate. These differences reflect NEAT library behavior (neat-python vs.\ TensorNEAT) rather than the hardware or parallelization strategy under study. Three categories of defaults differ between the two libraries: \emph{CPPN initial activation pools}, \emph{weight initialization and mutation parameters}, and \emph{aggregation and compatibility-threshold conventions}. Each library is the canonical companion to its respective implementation track; configuration files are released with the code (see Code and data availability) for full reproducibility and inspection of the differing library defaults. This confound affects the \emph{cross-implementation solve-rate comparison} only, not the structural-incompatibility claim, which is library-independent for the reasons detailed in Section~\ref{sec:caveats}.

\subsection{CartPole: A Positive Control Across Cost Regimes}
\label{sec:cartpole}

CartPole adds a reinforcement-learning control task (4 inputs; fitness is the normalized return over 5 episodes of CartPole-v1) whose per-evaluation cost is far higher than that of the Boolean and continuous benchmarks. Both implementations solve it at 100\% across depths 2--4 (Table~\ref{tab:cartpole}); the Baseline solves within a median of 4--6 generations and then early-stops, whereas JAX-ESHN solves at generation~1 but runs the full 100-generation budget. Because of this stopping asymmetry, total wall-clock times are not directly comparable, so we compare \emph{per-generation} compute cost, which is robust to how many generations each run executes.

\begin{table}[h]
\centering
\caption{CartPole (RL control) per-generation cost; both implementations on CPU ($n=30$, depths 2--4, Pop~150). Per-generation time isolates compute cost from the implementations' differing stopping behavior (Baseline early-stops at solve, median 4--6 generations; JAX-ESHN runs the full 100-generation budget). Both solve 100\% at every depth.}
\label{tab:cartpole}
\small
\begin{tabular*}{\textwidth}{@{\extracolsep{\fill}}l|c|rrr|r}
\toprule
& \textbf{Solve\%} & \textbf{D2 (s/gen)} & \textbf{D3 (s/gen)} & \textbf{D4 (s/gen)} & \textbf{D2$\to$D4} \\
\midrule
Baseline (CPU)  & 100 & $6.6 \pm 2.2$     & $56.2 \pm 32.5$   & $1{,}175.7 \pm 732.2$ & 177$\times$ \\
JAX-ESHN (CPU)  & 100 & $312.6 \pm 107.6$ & $476.7 \pm 118.5$ & $964.9 \pm 348.7$     & 3.1$\times$ \\
\bottomrule
\end{tabular*}
\end{table}

The per-generation cost reproduces the structural divergence seen on XOR, now on an RL task and entirely on CPU. The Baseline's per-generation cost grows 177$\times$ from D2 to D4 as quadtree traversal compounds with RL evaluation, while JAX-ESHN's grows only 3.1$\times$. The two curves cross: at D2 the Baseline is far cheaper per generation ($6.6$ vs.\ $312.6$~s, because JAX-ESHN pays a large fixed cost to evaluate the population's episodes through its vectorized pipeline), but by D4 the Baseline is the more expensive of the two ($1{,}176$ vs.\ $965$~s). This crossover mirrors the XOR result (Section~\ref{sec:post_jit}): wherever substrate depth is the dominant cost, the sequential quadtree's steep per-generation growth eventually overtakes the JAX pipeline, regardless of task type. The high variance in the Baseline's deeper-substrate cost (coefficient of variation ${\approx}0.6$) reflects the same bimodal early-solve-versus-slow-grind pattern noted for the other tasks.

\section{Discussion}
\label{sec:discussion}

Three implications follow from the structural diagnosis above. The quadtree's adaptivity (the \emph{topology uniqueness problem}) is intrinsic: removing it destroys solve rates and constrains the space of alternatives; preserving it leaves a con\-struc\-tion-over\-head plateau, expensive but predictable, and the resulting cost shifts from per-generation evolution to one-shot substrate construction. We discuss each below, then implementation caveats (Section~\ref{sec:caveats}) that bound the wall-clock comparison.

\subsection{Lessons from HSHG and Alternatives}

Running HSHG through the full evolutionary pipeline (XOR depths 2--3, Parity-3 depth 2, sine depth 2; Pop~50, 50 generations, 30 seeds per condition) makes the damage concrete: 0\% solve rate across all 120 runs, against the Baseline quadtree reference in Table~\ref{tab:hshg_task_performance}, which solves the Boolean tasks but not sine (quadtree mean best fitness 0.737 on sine, below the 0.95 threshold). The quadtree reference ran a 100-generation budget, so the comparison is not budget-matched; under a matched 50-generation cap the quadtree still solves Parity-3 in 18 of 30 seeds (60.0\%) against HSHG's zero.

\begin{table}[t]
\centering
\caption{Task performance through the full evolutionary pipeline: Quadtree vs.\ HSHG ($n=30$ per condition). HSHG ran at Pop~50 with a 50-generation cap; the quadtree reference ran at Pop~150 throughout with a 100-generation cap, because the harness used for these particular runs takes its population from the neat-python configuration rather than from the run parameter. The comparison is therefore unmatched on population as well as on budget, in the quadtree's favor. HSHG records a 0\% solve rate in every condition. This negative result supports the structural-incompatibility claim (Section~\ref{sec:introduction}) and is a third contribution, not the headline.}
\label{tab:hshg_task_performance}
\small
\begin{tabular*}{\textwidth}{@{\extracolsep{\fill}}lccc}
\toprule
\textbf{Task} & \textbf{Depth} & \textbf{Quadtree} & \textbf{HSHG} \\
\midrule
XOR      & 2 & 30/30 (100\%)        & 0/30 (0\%) \\
XOR      & 3 & 30/30 (100\%)        & 0/30 (0\%) \\
Parity-3 & 2 & 23/30 (76.7\%)       & 0/30 (0\%) \\
Sine     & 2 & 0/30 (mean fit 0.737) & 0/30 (0\%) \\
\bottomrule
\end{tabular*}
\end{table} Failures split into two modes: bloated substrates that prevent network construction (17--33\% of seeds), or functional networks plateauing far below threshold (Boolean tasks max 0.750: 3 of 4 patterns correctly classified, a majority vote that cannot represent XOR's nonlinear decision boundary; sine max 0.769 against a 0.95 solve threshold). Neither mode improves with continued evolution: fitness plateaus within the first 10 generations. Quadtree replacement requires rethinking how nodes are discovered, not swapping the data structure.

HSHG is not a one-off failure. It is the \emph{prototypical} static-position approach, and any alternative that pre-generates candidate positions and then filters by per-position variance (uniform grid sampling, hash-based spatial sweeps, learned continuous-density sampling over a fixed grid) inherits the same over-discovery pathology for the same structural reason. The pathology arises whenever the variance test is applied independently per position (Section~\ref{sec:hshg}). The space of viable alternatives that preserve adaptive sparsity is therefore narrow, which is itself a finding.

Two \emph{dynamic-shape} escapes lie outside the static-shape frameworks this paper targets; we do not rule them out, but neither is free. \emph{Ragged or jagged batching} avoids per-shape recompilation, yet replaces it with per-individual kernel dispatch whose overhead grows with the very per-CPPN cardinality variance that breaks \texttt{vmap}: the sequential population loop survives at the dispatch boundary, a different failure than the dynamic-recompilation route ruled out in Section~\ref{sec:optimizations}. \emph{Per-individual hand-written CUDA kernels} can accommodate variable shapes but reintroduce the sequential per-CPPN loop diagnosed in Sections~\ref{sec:introduction} and~\ref{sec:implementation}. The escape that works in practice does the opposite of embracing dynamic shapes: it keeps shapes static. EMR (Section~\ref{sec:emr}) shares one precomputed grid across the population, compiles once, and recovers adaptivity by masking rather than by per-genome subdivision.

\subsection{Construction-Overhead Plateau}

On the XOR benchmark, the construction-overhead plateau identified in Section~\ref{sec:results} (growth slowing from roughly $3\times$ per level to $1.2$--$1.4\times$ across the D5--D10 range) requires explanation. We did not isolate the exact cause. Plausible mechanisms include XLA graph saturation (a tracing-complexity ceiling once positions are numerous enough), fixed compilation overhead dominating variable graph-tracing cost at scale, XLA pattern-caching of repeated quadtree structures, and memory-bandwidth limits on compilation throughput; distinguishing them requires direct XLA-graph profiling.

Extended experiments show the plateau persists across the full D5--D10 range tested. This plateau contrasts sharply with the Baseline's continued exponential growth: while construction-overhead growth stays in the ${\sim}1.2$--$1.4\times$ band across the plateau range, Baseline total runtime continues scaling at $\approx 7\times$ per depth level (Table~\ref{tab:runtime_comparison}). This divergence explains why JAX-ESHN becomes competitive at deeper substrates despite its large upfront compilation cost.

\subsection{Where the Deep-Substrate Advantage Comes From}
\label{sec:post_jit}

On XOR, despite the construction overhead, JAX-ESHN becomes competitive at D7 because it solves within a few generations after compilation (Section~\ref{subsec:construction-cost}), while the Baseline (when it solves at all) requires extended evolution through its expensive per-generation substrate discovery. At D7 Pop~500 (Table~\ref{tab:runtime_comparison}), the Baseline averages $824 \pm 528$ minutes per attempt with only a 4.2\% solve rate; JAX-ESHN solves in $348 \pm 157$ minutes at 100\%. Accounting for retries, the expected wall-clock per successful solve is $\sim 327$ hours for the Baseline versus $\sim 5.8$ hours for JAX-ESHN: a $\sim 56\times$ effective advantage at this operating point. This ratio is retry-adjusted: it weighs JAX-ESHN's single reliable solve against the ${\sim}24$ attempts the Baseline needs at its 4.2\% success rate. On a matched 30-generation budget the comparison instead favors the Baseline: JAX-ESHN's projected total at D7 Pop~500 ($3{,}540$ minutes; Table~\ref{tab:runtime_comparison}) is more than $4\times$ the Baseline's. The same holds per generation once construction has amortized: subtracting construction overhead leaves JAX-ESHN at roughly $118$ minutes per generation at this operating point against the Baseline's ${\approx}27$, so amortization is where JAX-ESHN loses, not where it wins. It turns the comparison into a net win only by solving during the construction generation itself rather than exhausting the budget (Section~\ref{subsec:construction-cost}). The deep-substrate advantage is real but narrow, confined to the regime where the Baseline's solve rate has collapsed; JAX-ESHN is best read as a diagnostic reimplementation, not a general-purpose accelerator. JAX's persistent compilation cache does not alleviate the compilation burden: each CPPN's unique substrate topology yields distinct tensor shapes within the population, requiring separate compilation per CPPN.

The regime in which this helps is narrow. Because the post-construction per-generation cost runs against JAX-ESHN at these depths, extended search compounds the deficit rather than amortizing it away: the more generations a problem needs, the worse the trade becomes. The advantage survives only where a deep substrate is required and a solution appears almost immediately, which is what XOR provides here.

We therefore recommend the Baseline (CPU) for shallow substrates (D1--D5), and for extended evolution on this XOR workload, where its per-generation cost is the lower of the two at every depth measured. JAX-ESHN is preferable at D6+ where the Baseline's solve rate has collapsed, and there it wins on reliability rather than on throughput. Which implementation is cheaper per generation is campaign-dependent: on the CPU multi-benchmark tasks the comparison runs the other way (Section~\ref{sec:multi_benchmark}).

\subsection{Implementation Comparison Caveats}
\label{sec:caveats}

We separate two claims in this paper because they carry different cross-im\-ple\-men\-ta\-tion assumptions. The \emph{structural-incompatibility claim} (Sections~\ref{sec:introduction}, \ref{sec:optimizations}, \ref{sec:hshg}) is library-independent: it follows from the variable-cardinality output of quadtree subdivision, which holds regardless of whether neat-python or TensorNEAT drives the CPPN evolution upstream. By contrast, the \emph{wall-clock and solve-rate comparison} (Section~\ref{sec:results}) is library-dependent: differences in CPPN initialization, parameter defaults, activation pools, and speciation thresholds (cataloged in Section~\ref{sec:multi_benchmark}) propagate into solve-rate and per-generation timing differences. Normalizing the two implementations to identical NEAT-library configurations is not possible in practice: TensorNEAT exists precisely to expose NEAT to JAX/XLA, so there is no GPU-accelerated neat-python comparison to run; conversely, porting TensorNEAT's configuration onto neat-python would defeat the GPU comparison this paper is built around. We therefore report both implementations with the library confound explicit and tag each quantitative claim: structural claims stand library-independent; wall-clock comparisons require this caveat.

A second consideration concerns hardware and the JAX backend, because the two campaigns differ. The XOR scaling study (Section~\ref{sec:results}) runs JAX-ESHN on an NVIDIA RTX~2080~Ti GPU against the Baseline on an Apple~M4~Max CPU, so its wall-clock comparison spans both a different implementation \emph{and} different hardware. We measured GPU utilization during that study and found it low and erratic (per-configuration means in the timing reruns span ${\sim}0$--$65\%$ with no consistent trend in population size): the device is never saturated, consistent with only within-CPPN operations vectorizing while the population loop stays sequential, though we do not lean on this number as evidence (utilization sampling conflates compilation and evaluation phases). The multi-benchmark campaign (Section~\ref{sec:multi_benchmark}) instead runs JAX-ESHN on JAX's \emph{CPU} backend, matching the Baseline's CPU execution; it therefore controls for hardware, removing the GPU-vs-CPU dimension entirely and showing that the steep-versus-shallow scaling divergence persists on CPU alone. A residual difference remains even there: the two CPU runs used different machines (Apple Silicon for the Baseline, an x86 host for JAX-ESHN), so absolute per-generation times are not strictly comparable across implementations; the depth-\emph{scaling} ratios and solve rates, computed within each implementation, are the robust quantities. The structural-incompatibility claim is unaffected by backend, because it concerns the variable-shape output of quadtree subdivision under static-shape compilation, which is identical on CPU and GPU XLA backends.

A third caveat concerns connectivity conventions, on which the two implementations differ in two ways. First, the feedforward test: the Baseline admits a connection only when $s.y < \ell.y$, whereas JAX-ESHN admits $s.y \le \ell.y$ and can therefore also connect two discovered nodes that share a layer coordinate. Because inputs sit at $y=-1$, outputs at $y=+1$, and quadtree subdivision reaches only the interior, this can affect hidden-to-hidden connections alone; we did not measure how often it does. Second, both implementations run \texttt{iteration\_level}~1 here, but higher iteration levels, which the Baseline supports and which its XOR scaling campaign does exercise (Section~\ref{sec:results}), require iterative propagation whose step-to-step dependencies resist \texttt{vmap} batching, so no JAX-ESHN campaign uses them. Both differences belong to the same class as the library defaults cataloged above: they bear on the wall-clock and solve-rate comparison, not on the structural claim.

\subsection{From Diagnosis to Solution: EMR-HyperNEAT}
\label{sec:emr}

The structural barrier characterized here is not the end of the story. Eager Multi-Resolution HyperNEAT (EMR-HyperNEAT)~\cite{claret2026emr} resolves it by inverting the order of discovery and evaluation. Where ES-HyperNEAT subdivides \emph{lazily} (\texttt{subdivide if variance} ${>}\,\theta$, the data-dependent recursion that produces the per-CPPN variable shapes diagnosed here), EMR evaluates \emph{eagerly}: it precomputes a single static multi-resolution grid containing all positions to depth $D$, queries every position for the whole population in one batched \texttt{vmap}, computes variance bottom-up, then applies the variance threshold through a top-down mask. Because the grid is fixed and shared across the population, every tensor shape is static, so the population compiles \emph{once} rather than once per CPPN: the recompilation barrier of Section~\ref{sec:optimizations} disappears. The reformulation reduces the sequential $O(4^D)$ traversal to $O(4^D/P)$ across $P$ cores; on XOR (population~1000) EMR overtakes ES-HyperNEAT at depth~4 and reaches a $34\times$ per-generation speedup at depth~7 (EMR on GPU against ES-HyperNEAT on CPU), with the 30-generation total-runtime ratio approaching ${\sim}100\times$ (IQR median) once JIT compilation amortizes, while matching solution quality (both reach $0.99{+}$ fitness; EMR solves $100\%$ of hidden-to-hidden configurations in that comparison).

EMR is \emph{not} the HSHG approach that fails in Section~\ref{sec:hshg}, even though both pre-generate positions. HSHG applies the variance test to each pre-generated position \emph{independently}, so adjacent positions that overlap one CPPN feature each pass the test and over-discover. EMR instead filters \emph{hierarchically}: its mask gates each parent's four children as a block (a child is active only if its parent's variance exceeds the threshold), preserving the quadtree's grouped-pruning logic. This is exactly the distinction Section~\ref{sec:hshg} draws: a fixed grid is viable only when the variance criterion is applied with the quadtree's parent--child structure, not position-by-position. Under matched thresholds EMR provably discovers a superset of the quadtree's positions~\cite{claret2026emr}, never fewer: the gap is one of evaluation, not of the keep-rule. EMR computes variance bottom-up over the whole pre-generated grid before masking, whereas the lazy quadtree stops subdividing a region before examining what lies beneath it, so under variance non-monotonicity (clearest at \texttt{initial\_depth}~0, where a low root estimate can halt the quadtree outright) EMR retains structure the quadtree prunes unexamined. Because the mask is still parent-gated, this superset is marginal, not the order-of-magnitude over-discovery of HSHG; the released EMR runs in fact use a more permissive variance threshold than the Baseline yet stay non-bloated for the same structural reason, keeping the theorem's matched-threshold premise distinct from the empirical configuration. In the terms of this paper, EMR is an eager-evaluation, adaptivity-preserving substrate-discovery method: the redesign of discovery that the HSHG failure shows is required.

\section{Conclusion}
\label{sec:conclusion}

ES-HyperNEAT's adaptive quadtree subdivision is structurally incompatible with static-shape compilation frameworks (JAX \texttt{vmap}/XLA and analogues). This answers our three research questions. Batched optimizations cannot overcome the quadtree's sequential substrate discovery (RQ1): within-CPPN batching plateaus near $1.7\times$ (XOR, depth~2) because population-level vectorization stays blocked. For RQ2, JAX-ESHN trades shallow-depth wall-clock for deep-substrate reliability, reaching near-100\% solve rates at depths where the Baseline collapses to 4.2\%, at the price of construction overhead that dominates total runtime. The scalability limit is construction-bound, not search-bound (RQ3). Once a deep substrate is built, search is nearly trivial, and construction cost itself plateaus beyond depth 5. The quadtree's per-CPPN refinement is the essential mechanism that produces sparse topologies, and it cannot simply be removed: replacing it with HSHG failed to solve any task. The discovery approach itself must change, and any replacement must satisfy three structural constraints: tensor shapes static and shared across the population, variance filtering applied through the quadtree's parent-gated hierarchy rather than independently per position, and adaptive sparsity preserved under that filtering. Dynamic-shape GPU runtimes remain a theoretical escape that this work does not rule out, but no production framework currently provides this for evolutionary loops. A different escape does work: eagerly evaluating a static multi-resolution grid with hierarchical, parent-gated variance filtering removes the recompilation barrier while preserving adaptive sparsity (Section~\ref{sec:emr}).

Validation across five benchmarks (XOR, Parity-3, circle, sine, CartPole) spanning Boolean, continuous, and control task types shows the scaling divergence is not task-specific: Baseline runtime grows steeply as solve rates collapse, while JAX-ESHN runtime grows far more slowly, and a CPU-vs-CPU control rules out GPU hardware as the cause.

\paragraph{Future work.} Testing on higher-dimensional problems such as MNIST would strengthen the generality claim. We did not test multithreaded JAX on CPU, which would not change the vectorization conclusion but could narrow the runtime gap at shallow depths.

\paragraph{Code and data availability.} The JAX-ESHN implementation, the PUREPLES Baseline harness, the benchmark driver and configuration files for the five tasks, and the per-seed raw result files (timing, fitness, topology, and GPU-utilization metrics) for the JAX-ESHN scaling, multi-benchmark, CartPole, and HSHG campaigns are released in a public GitHub repository,\footnote{\url{https://github.com/RomainClaret/jax-es-hyperneat}} together with analysis and verification scripts that recompute the scaling and multi-benchmark tables from the raw files; the raw measurement archive is also deposited on Zenodo (\url{https://doi.org/10.5281/zenodo.21761118}). Two measurements are outside that release: the optimization speedups of Table~\ref{tab:optimizations} and the substrate-discovery probe of Section~\ref{sec:hshg}, both recorded during development without surviving per-run files.

\bibliographystyle{splncs04}
\bibliography{references}

\appendix
\section{Reproducibility Details}
\label{sec:repro}

This appendix records the hardware, software stack, hyperparameters, seeds, and data provenance for every reported number. The two experimental campaigns differ in backend and sampling, so we separate them.

\paragraph{Hardware and backend.} The Baseline (PUREPLES + neat-python) ran on an Apple~M4~Max CPU. JAX-ESHN ran on NVIDIA RTX~2080~Ti hardware (11~GB): on the \emph{GPU} backend for the XOR depth/population scaling study (Section~\ref{sec:results}), and on the \emph{CPU} backend for the multi-benchmark campaign (Section~\ref{sec:multi_benchmark}). The scaling study spans the campaign period rather than a single host image, and the released per-run records report driver versions 550.120 to 580.126.09 and CUDA 12.4 to 13.0; the multi-benchmark and CartPole runs are all driver 580.126.09 with CUDA~13.0. The released repository pins the package versions used (JAX, TensorNEAT, neat-python~0.92, PUREPLES, NumPy, Python) in its packaging metadata and continuous-integration configuration. Per-benchmark hyperparameters are listed in Table~\ref{tab:repro_config}.

\begin{table}[h]
\centering
\caption{Per-benchmark configuration. $D$~denotes \texttt{max\_depth} at \texttt{initial\_depth}~$=0$. Shared across all: \texttt{division\_threshold}~$=0.5$, \texttt{variance\_threshold}~$=0.03$, \texttt{iteration\_level}~$=1$, \texttt{band\_threshold}~$=0.3$, substrate activation sigmoid, CPPN activation pool $\{$tanh, sin, gauss$\}$. \texttt{max\_weight} is $5.0$ (PUREPLES) and $8.0$ (JAX-ESHN), one of the implementation-default differences discussed in Section~\ref{sec:caveats}. Paired cell values read Baseline/JAX-ESHN. The XOR campaigns used different fitness thresholds and disjoint seed sets, so the Baseline was held to the stricter bar; re-scoring every Baseline run at $0.98$ changes a single D3 trial ($62.5\% \rightarrow 66.7\%$) and leaves every other depth cell in Table~\ref{tab:runtime_comparison} unchanged, D7 included. Seeds control the JAX-ESHN runs throughout. On the Baseline side they control the XOR scaling campaign, but not the multi-benchmark and CartPole campaigns: that harness seeds NumPy's generator whereas neat-python~0.92 draws from Python's \texttt{random} module, so those Baseline seeds label runs without determining them.}
\label{tab:repro_config}
\small
\begin{tabular*}{\textwidth}{@{\extracolsep{\fill}}l|ccc}
\toprule
\textbf{Setting} & \textbf{XOR (scaling)} & \textbf{Parity-3 / circle / sine} & \textbf{CartPole} \\
\midrule
Depths $D$            & 1--7 (8--10 expl.) & 2--4              & 2--4 \\
Population            & 50--1000 (8/10 sizes)   & 150               & 150 \\
Generations          & 10 (timing) / 30/300 (solve) & 100        & 100 \\
Seeds ($n$)          & 3 (42,123,456 / 42,43,44) & 30 (0--29)        & 30 (0--29) \\
Fitness threshold    & 0.99 / 0.98        & 0.975 (circle, Parity-3) & 0.95 \\
                     &                    & 0.95 (sine)       & \\
JAX backend          & GPU                & CPU               & CPU \\
\bottomrule
\end{tabular*}
\end{table}

\paragraph{Data provenance.} The JAX-ESHN timing entries in Table~\ref{tab:runtime_comparison} and Table~\ref{tab:jit_by_pop} draw on 10-generation forced reruns at depths D1--D7 (seeds 42/43/44), and the JAX-ESHN solve rates on the 300-generation solve campaign. The 30-generation \emph{Total} in Table~\ref{tab:runtime_comparison} is a projection (construction overhead plus $29\times$ the measured post-construction per-generation time, generation~1 being inside construction overhead), not a measured 30-generation wall-clock. The D8--D10 points in Figure~\ref{fig:baseline_iqr} and Section~\ref{subsec:exploratory_deep} are measured at Pop~1000 in a separate 300-generation campaign ($n=3$). The line is intentionally broken between D7 and D8 because the two campaigns measure construction overhead differently. The Baseline ANOVA statistics aggregate a 456-trial XOR campaign under a 30-generation budget (D1--D6: 6 depths $\times$ 8 populations $\times$ 3 iteration levels $\times$ 3 replications; D7 at iteration level~1); the Baseline solve rates in Table~\ref{tab:runtime_comparison} use its iteration-level-1 subset (24 trials per depth). The multi-benchmark tables (Tables~\ref{tab:multi_benchmark}--\ref{tab:cartpole}) report 100-generation runs at $n=30$. The CartPole comparison uses per-generation cost because PUREPLES early-stops at solve (median 4--6 generations) while JAX-ESHN runs the full 100-generation budget.

\end{document}